\documentclass[runningheads]{llncs}
\usepackage{graphicx}
\usepackage{amssymb}
\usepackage{array}
\usepackage{svg}
\usepackage{bbold}
\usepackage{booktabs}
\usepackage{multirow}
\usepackage{mathrsfs} 
\begin{document}
\title{COIN: Counterfactual inpainting for weakly supervised semantic segmentation for medical images
}
%
%
\author{Dmytro Shvetsov\inst{1}\orcidID{0009-0005-7512-9604} \and
Joonas Ariva\inst{1,2,3}\orcidID{0009-0003-7171-7359} \and
Marharyta Domnich\inst{1}\orcidID{0000-0001-5414-6089} \and
Raul  Vicente\inst{1}\orcidID{0000-0002-2497-0007} \and
Dmytro Fishman\inst{1,2,3}\orcidID{0000-0002-4644-8893}}
\authorrunning{Shvetsov et al.}
\titlerunning{Counterfactual inpainting for weakly supervised segmentation}
%
\institute{Institute of Computer Science, University of Tartu, Tartu, Estonia \email{marharyta.domnich@ut.ee}  \and Better Medicine, Tartu, Estonia  \and STACC, Tartu, Estonia}
\maketitle              
\begin{abstract}
Deep learning is dramatically transforming the field of medical imaging and radiology, enabling the identification of pathologies in medical images, including computed tomography (CT) and X-ray scans. However, the performance of deep learning models, particularly in segmentation tasks, is often limited by the need for extensive annotated datasets. To address this challenge, the capabilities of weakly supervised semantic segmentation are explored through the lens of Explainable AI and the generation of counterfactual explanations. The scope of this research is development of a novel counterfactual inpainting approach (COIN) that flips the predicted classification label from abnormal to normal by using a generative model. For instance, if the classifier deems an input medical image X as abnormal, indicating the presence of a pathology, the generative model aims to inpaint the abnormal region, thus reversing the classifier’s original prediction label. The approach enables us to produce precise segmentations for pathologies without depending on pre-existing segmentation masks. Crucially, image-level labels are utilized, which are substantially easier to acquire than creating detailed segmentation masks. The effectiveness of the method is demonstrated by segmenting synthetic targets and actual kidney tumors from CT images acquired from Tartu University Hospital in Estonia. The findings indicate that COIN greatly surpasses established attribution methods, such as RISE, ScoreCAM, and LayerCAM, as well as an alternative counterfactual explanation method introduced by Singla et al. This evidence suggests that COIN is a promising approach for semantic segmentation of tumors in CT images, and presents a step forward in making deep learning applications more accessible and effective in healthcare, where annotated data is scarce.

\keywords{Explainable AI \and Counterfactual explanations  \and GANs \and Semantic Segmentation \and Medical Imaging \and CT scans \and Kidney Tumour.}
\end{abstract}

\section{Introduction}
Deep learning is revolutionizing the field of medical imaging \cite{burton2019using,musen2021clinical} and radiology \cite{gidde2021validation}, offering the potential to aid radiologists by triaging incoming patients and detecting various pathologies from medical images like computed tomography (CT) or X-ray scans. However, the accuracy of deep learning models for medical images hinges on access to substantial annotated datasets \cite{cui2020unified}. Collecting such datasets is hard for two reasons: 1) labeling medical images accurately requires the knowledge of trained medical professionals such as radiologists, 2) accurate manual image labeling is very labor intensive. These problems are further amplified by the limited access to medical image datasets due to data protection laws \cite{chaddad2023survey}. 

Considering these challenges, there is a pressing need for methods that can automate or simplify the manual data labelling process. Dense pixel-level annotations, though highly informative, are particularly time-consuming to create \cite{tajbakhsh2020embracing}. In contrast, image-level annotations, which indicate the presence or absence of certain organs or pathologies, are more feasible to obtain and can be derived from accompanying radiology reports. This scenario raises an intriguing question: Can one generate detailed pixel labels based solely on image-level labels?

This task falls under weakly supervised semantic segmentation (WSSS) and it is often tackled with methods known from Explainable AI (XAI) \cite{ahn2018learning,chen2020weakly,chen2022c,shen2023survey}. In computer vision, the XAI’s task is usually to explain a “black box” classifier by highlighting the most important regions of the images for classifiers decisions. The principle of generating saliency maps from a classifier is partly transferable over to the task of WSSS as the saliency maps can be seen as segmentation masks. However, as noted in \cite{ghassemi2021false}, saliency maps, while visually intuitive, often blend useful and non-useful information, making it hard to identify the specific image features that are important for model's decisions. This makes it difficult to generate precise segmentations from the saliency maps.

Counterfactual explanations recently emerged that seeks for minimal change in input to flip the decision output of classifier \cite{wachter2017counterfactual,miller2021contrastive}. This approach aims not only to flip the decision of a model in adversarial attack manner, but to ensure that modifications are meaningful and interpretable in a real-world context as highlighted in \cite{zemni_octet_2023,jeanneret_adversarial_2023}. First trained counterfactual explanation models with usage of conditional Generative Adversarial Network (cGAN) \cite{singla_explaining_2022} showed great promise in providing insights into decision-making processes of classifiers and uncovering potential biases or failure modes. Building upon this work, this research explores the potential of adapting the counterfactual explanation framework for the domain of WSSS. It is argued that the difference between original input and its respective counterfactual image can serve as implicit segmentation masks, while also revealing critical features for classification decisions. 

In this context, this study extends the application of the generative counterfactual explanation method to facilitate the generation of segmentation labels. The original methodology and architecture are refined to better suit the needs of WSSS, thereby eliminating the dependence on pre-existing segmentation masks for training. This approach allows for the development of a more efficient, weakly supervised learning framework, enhancing the precision of segmentation outcomes. The method is tested by segmenting kidney tumors from CT images. The adapted counterfactual pipeline efficiently produces accurate segmentation labels from straightforward classification models. This innovation is substantiated through comprehensive testing and validation on a synthetic dataset. Furthermore, the novel counterfactual pipeline is compared against established attribution methods — RISE \cite{petsiuk_rise_2018}, Score-CAM \cite{wang_score-cam_2020}, and LayerCAM \cite{jiang_layercam_2021} — to affirm its enhanced capability and practical applicability in generating segmentation labels under weak supervision.

\section{Related Works}

\subsection{Weakly supervised semantic segmentation}

WSSS has gathered a lot of attention as the idea of segmentation masks from image-level labels makes it very cost-effective. WSSS methods often utilize class activation maps (CAMs). In CAM methods the saliency maps are generated from classifier model and input image to indicate the most important image regions for the classifiers decision for each class. Frequently used CAM methods include GradCAM, ScoreCAM and LayerCAM \cite{selvaraju2017grad,wang_score-cam_2020,jiang_layercam_2021}. CAM based WSSS methods consist of the following steps: 1) Training of the classifier 2) Extracting saliency maps with a CAM method  3) Refining the saliency maps by postprocessing. Additionally, sometimes these refined maps are used to train a segmentation model and then the model is used to acquire the final segmentations. 

However, CAM methods are not ideal for WSSS and are not without issues. One of the major problem with CAMs is that they highlight only the most discriminative regions of the image and not the full object that represents the class. This can lead to poor segmentation performance. Secondly, the resolution of saliency maps from CAMs is tied to the resolution of the activation maps from the classifier model. Using the activations from deepest layers of the model usually yields semantically more representative saliency maps. At the same time these activations have low resolution which creates low resolution saliency maps. Moreover, saliency maps can remain unchanged even when the underlying model predictions are significantly affected by adversarial attacks as pointed in \cite{ghassemi2021false}, raising questions about their reliability as explanations. 

The potential way to overcome these issues in WSSS is to use counterfactual explanations instead of CAMs.

\subsection{Counterfactual explanations}
Counterfactual explanation is model-agnostic instance-based method that answers the question "What is the minimum input change that leads to the flip of the prediction outcome?".
Originating from the fields of cognitive science \cite{byrne2019counterfactuals}, psychology \cite{keil2006explanation}, and causality research \cite{pearl2019seven}, counterfactual explanations have been explored as a way to understand AI's decision-making processes.  Wachter et al. \cite{wachter2017counterfactual}, introduced a formal counterfactual optimization function for generating explanations in continuous data, marking a significant milestone in XAI. Furthermore, Tim Miller's insights from social sciences \cite{miller2019explanation} highlighted the criticality of contrastive explanations for human reasoning and decision-making processes and the necessity of counterfactual explanations for XAI. Extensive surveys on counterfactual explanations by Guidotti et al. \cite{guidotti2022counterfactual} and Karimi et al. \cite{karimi2022survey} showed big variety of counterfactual explanation frameworks primarily focused on discrete data. These frameworks either solve optimization problems or employ heuristics search strategies to find counterfactual explanations. In image domain a notable advancement introduced Akula et al.  \cite{akula2022cx}, who highlighted the importance of Theory of Mind for Explainability and developed a pipeline for generating counterfactuals in images by identifying and modifying minimal semantic-level features, such as altering the stripes on a zebra.

The application of Generative Adversarial Networks (GANs) for creating counterfactual explanation has gained increasing attention, demonstrating the versatility across a variety of domains. For instance, Kenny et al. \cite{kenny2021generating} PIECE method demonstrated this on MNIST data by altering exceptional image features to generate plausible counterfactuals for black-box CNN classifiers. In the domain of autonomous driving, a field where explainability is crucial due to the safety-critical nature of its applications, Zemni et al.  \cite{zemni_octet_2023} proposed an object-centric framework for coutnerfactual generation. Their method was specifically designed for images with many objects, such as urban scenes common in autonomous driving. By encoding the query image into a structured latent space, this approach facilitates object-level manipulations, making it highly suitable for complex scenes. The method was tested on counterfactual explanation benchmarks for driving scenes, demonstrating its capability to adapt beyond classification to explain semantic segmentation models. Jeanneret et al. \cite{jeanneret_adversarial_2023} focused on transforming adversarial attacks into semantically meaningful perturbations for facial expression data. Their work hypothesized that Denoising Diffusion Probabilistic Models could regularize adversarial attacks to generate actionable and understandable image modifications, such as making sad people happy. Bischof et al. \cite{bischof_counterfactual_2023} proposed a unified framework leveraging image-to-image translation GANs to address interpretability and robustness in neural image classifiers. This framework was designed and assessed on two specific applications: a semantic segmentation task for concrete cracks and a fruit defects detection problem. Through this, they produced saliency maps for interpretability and demonstrated improved model robustness against adversarial attacks.

In the medical field, counterfactual explanations have shown great promise for diagnostic purposes, particularly in analyzing chest X-ray images. The work by Atad et al., \cite{atad2022chexplaining} employed a StyleGAN-based approach, StyleEx, to manipulate specific latent directions in chest X-ray images. They demonstrated the use of counterfactual explanations in analyzing chest X-ray images helping to identify the patterns that models rely on for diagnoses, which was clinically evaluated with radiologists. Similarly Singla et al. \cite{singla_explaining_2022} used counterfactual explanation generation for chest X-rays images to explain the decision-making processes of image classifiers. They trained a  generative model capable of producing images that would lead to a different classification by the original model by preserving original context from an instance. Their generative model was trained and validated on chest X-ray images, with a human-grounded study confirming the usefulness of the generated explanations in a medical context. 

While these works have illustrated the prominent capabilities of counterfactual explanations across various tasks, their application from the perspective of WSSS remains unexplored. Most methods have relied on segmentation masks within GAN training, which may not be available in this context. COIN aims to bridge this gap by adapting the counterfactual approach for segmentation purposes, particularly focusing on the work of  Singla et al.'s \cite{singla_explaining_2022} methodology and adapting it to generate segmentation labels without using pre-existing masks.

\section{Counterfactual approach for WSSS}

The counterfactual inpainting approach is introduced for producing semantic segmentation masks. Counterfactual approach originates from
Singla et al. \cite{singla_explaining_2022} with significantly enhanced architecture, adding perturbation-based generator, skip connections, another loss measure and discarding the usage of segmentation masks to make it more viable for
WSSS.

\subsection{Method formulation} 
\begin{figure}[!ht]
    \centering
    \includegraphics[width=1\linewidth]{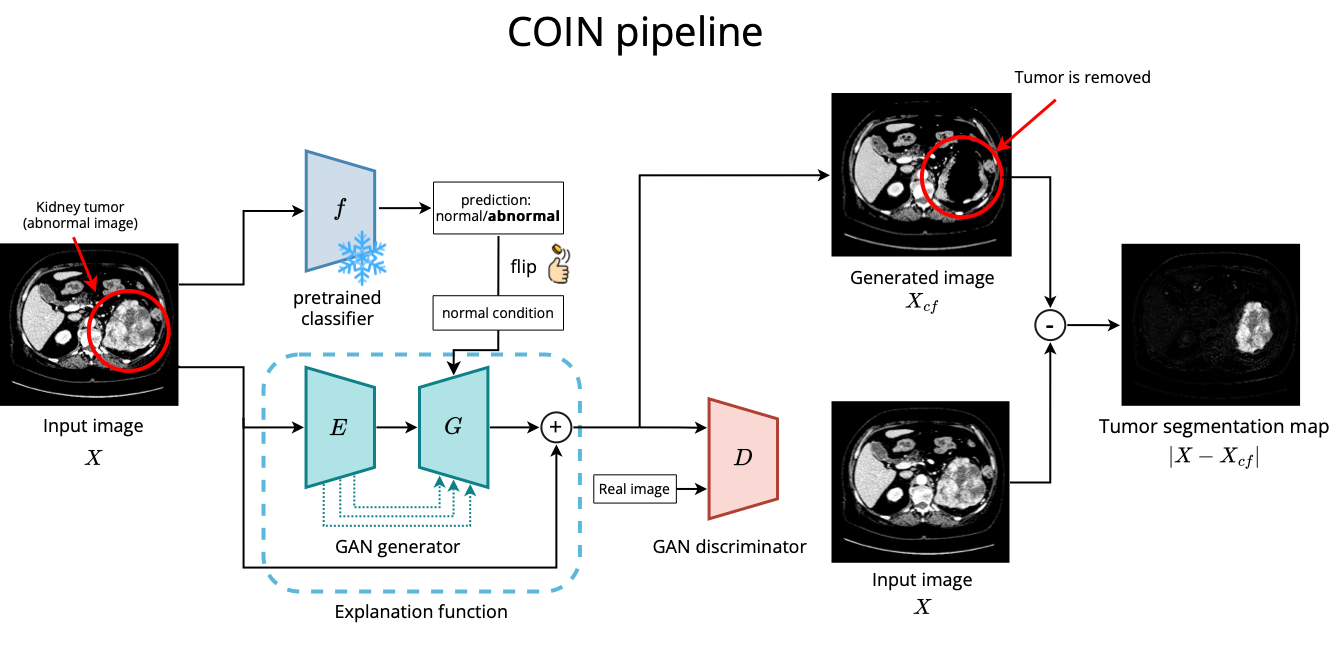}
    \caption{Overview of the proposed counterfactual inpainting (COIN) pipeline. Given the input image $X$ and black-box classifier $f$ that produces a classification label, the image-to-image model (GAN) generates a counterfactual image $X_{cf}$ with $y = 0$. If $X$ is \textbf{abnormal}, it is expected that $X_{cf}$ no longer contains the abormal part of the input image. Computing the absolute difference of the original image $X$ and counterfactual image $X_{cf}$ results in a weak tumor segmentation map. While training the pipeline, only GAN weights are updated. Classifier predictions are used for classifier consistency loss calculation.}
    \label{fig:coin_pipeline}
\end{figure}

Our method is defined in the case of a binary classification. Let's denote a black-box classifier as $f$ with assumption that it is differentiable with access to its output value and gradient. The pre-trained binary classifier $f$ accepts an input image $X$ and outputs whether an image $X$ is \textbf{normal} ($y=0$) if $f(X)<t$ or \textbf{abnormal} ($y=1$) if $f(X) \geq t$, where $t$ is a threshold used to binarize the prediction output of $f$. In Singla et al. \cite{singla_explaining_2022}, the generative model (cGAN) accepts an input image $X$ and a tweakable parameter $\delta$, such that the counterfactual image $X_{cf} = \mathscr{E}(X, \delta)$  flips the prediction class $y$ and $\mathscr{E}(\cdot)$ is used as an explanation function. Specifically, if the classifier predicts \textbf{abnormal} for an image $X$, the generator aims to inpaint or remove the abnormal region, effectively \textbf{flipping} the classifier's prediction. Similarly, if the classifier predicts \textbf{normal}, the generated counterfactual image should add the \textbf{abnormal }region, \textbf{flipping} the classifier's prediction. In contrast, this research argues that only the inpainting case should be considered for WSSS purposes. Our generative model predicts a \textbf{normal} counterfactual image $X_{cf}$ only when the classifier deems the input image $X$ as \textbf{abnormal}. Subsequently, the absolute difference between the counterfactual image $X_{cf}$ and the original image $X$ serves as a weak segmentation label. The overview of the COIN method is depicted in Figure \ref{fig:coin_pipeline}.

\subsection{Image generation architecture}
The SNGAN \cite{miyato_spectral_2018}  architecture is adapted for the generator and discriminator networks. The original generative model (cGAN) consists of an encoder $E(X)=z$ and a decoder $G(z, \delta)=X_{cf}$. Consequently, the explanation function can be expressed as $\mathscr{E}(X, \delta)=G(E(X),\delta)$. The encoder transforms an input image $X$ into a latent representation $z$, which is then passed into the decoder along with a condition label $\delta$ to produce a counterfactual image $X_{cf}$. Unlike the approach by Singla et al. \cite{singla_explaining_2022}, where the parameter $\delta$ is utilized to provide unique offsets by discretizing the $[0;1]$ range into $N=10$ equal bins — each corresponding to a unique condition vector $c$ — COIN simplifies the architecture to handle only one condition ($c_{inp}$ - to inpaint or remove the abnormal region). The model is trained as a simple GAN that produces counterfactual image $X_{cf}$ where $f(X_{cf}) < t$. In this case, the flipping happens only in one direction when the $X$ is \textbf{abnormal} and $X_{cf}$ becomes \textbf{normal}, removing the area that affects the classifier's prediction. This approach addresses the challenge of needing a condition-balanced training set for the cGAN by reducing the complexity and data requirements. The benefits of decreasing number of condition is showcased in Appendix A Table~\ref{tab:iterative_impr}.

Moreover, COIN distinguishes itself by implementing a perturbation-based image generation within the GAN framework. The generator architecture is unique with slight modification of having a residual connection of the model's input to the outputs. Instead of regenerating the complete image in the decoder from the latent variable, the generator is trained to produce only the perturbation map which is fused into the input image. Therefore, the explanation function is modified such that $\mathscr{E}(X, \delta)=G(E(X),\delta)+X$. This technique contrasts with conventional counterfactual generation models \cite{bischof_counterfactual_2023,jeanneret_adversarial_2023,zemni_octet_2023}, which typically reconstruct the entire image. While full reconstruction can be effective in some contexts, it often introduces artifacts — unintended alterations that can skew the classification model's interpretation and analysis. The impact of perturbation-based architecture is illustrated in Appendix A Figure~\ref{fig:orig-vs-ptb-based-vis} and Table~\ref{tab:iterative_impr}.

Additionally, in contrast to reported Singla et al. architecture, the skip-connections are integrated into the encoder-decoder model \cite{ronneberger_u-net_2015} together with perturbation-based approach. This enhancement facilitates the generation of more accurate perturbations, thereby improving reconstruction quality and preserving the original image details. The impact of skip-connections is showcased in Appendix A Figure~\ref{fig:skip-conn-vis} and Table~\ref{tab:iterative_impr}. This approach is beneficial for WSSS as it ensures the generation of precise, artifact-free counterfactual explanations.

\subsection{Loss function for training GAN}

The same loss functions are inherited as described by Singla et al \cite{singla_explaining_2022}, and introduce an additional Total-Variation loss \cite{javanmardi_unsupervised_2018} to enforce smoothness in the generated images.  The complete objective function for the counterfactual inpainting pipeline is given as follows:
\begin{equation}
\min_{E,G}\max_{D} \;( \lambda_{GAN}{L}_{\mathrm{GAN}} + \lambda_{f}{\mathcal L}_{f} + \lambda_{idt}{\mathcal{L}}_{\mathrm{idt}} + \lambda_{tv}{\mathcal{L}}_{\mathrm{tv}}),
\end{equation}
where $E$, $G$ and $D$ is the Encoder, Decoder and Discriminator of GAN; $f$ is a pre-trained classifier; ${L}_{\mathrm{GAN}}$ is a data consistency loss term; ${\mathcal L}_{f}$ is a classification model consistency loss term; ${\mathcal{L}}_{\mathrm{idt}}$  is a domain-aware self-consistency loss term, and ${\mathcal{L}}_{\mathrm{tv}}$ is a Total-variation loss term; $\lambda_{GAN}$, $\lambda_{f}$, $\lambda_{idt}$, $\lambda_{tv}$ are respective hyper-parameters to configure contribution of each term. 

\subsubsection{Data consistency loss term.}
Generated images should look similar to the images in the training dataset. For GAN, the two networks are trained: generator, which consists of encoder $E(\cdot)$ and decoder $G(\cdot)$, and discriminator $D(\cdot)$ and compute binary cross-entropy loss on the real/fake labels. 
\begin{equation}
\mathcal{L}_{\mathrm{GAN}}=\mathbb{E}_{X \sim P(X)}\left[\log\left(D(X)\right)\right]\ +\ \mathbb{E}_{X \sim P(X_{cf})}\bigl[\log(1-D(E(G(X))\bigr],
\end{equation}
where $P(X)$ and $P(X_{cf})$ denote distributions of real and generated images respectively. 

\subsubsection{Classification model consistency loss term.}
A GAN should generate the counterfactual images that influence the classifier predictions in the desired manner. The general formulation of the objective function computes the KL divergence between the predicted and expected distributions of probabilities. In the proposed reformulation for the inpainting pipeline, the condition-aware loss simplifies to:
\begin{equation}
{\mathcal L}_{f} =D_{\mathrm{KL}}(f(\mathscr{E}({X}))\ ||\ 0),
\end{equation}
where $\mathscr{E}({X})$ is a counterfactual explanation derived from an input image X.
\subsubsection{Domain-aware self-consistency loss term.}

Similarly to Singla et al \cite{singla_explaining_2022}, the main idea behind the objective function is to let the model learn cyclically consistent counterfactual images when applying a series of counterfactual generations. One cycle of generations is computed to produce $\mathscr{E}(X)$ and $\mathscr{E}(\mathscr{E}(X))$, given that $\mathscr{E}(\cdot)$ - is an explanation function. Both counterfactual images should retain as much details as possible from the input image perturbing it only if it is abnormal. Additionally, Singla et al. \cite{singla_explaining_2022} uses segmentation masks to enforce local consistency over the foreground pixels of different segmentation labels present in the image. However, this research employs a simpler supervision, not requiring segmentation masks, to compute the reconstruction loss over the whole image instead of local regions. The domain-aware self-consistency loss for the proposed counterfactual inpainting pipeline is given by:
\begin{equation}
{\mathcal{L}}_{\mathrm{idt}}={\mathcal{L}}_{\mathrm{1}}(X,\mathscr{E}(X))\ +\ {\mathcal{L}}_{\mathrm{1}}(X,\mathscr{E}(\mathscr{E}(X))),
\end{equation}
where
\begin{equation}
{\mathcal{L}}_{\mathrm{1}}(X,X^{\prime})={\frac{||X-X^{\prime}||_1}{HW}},
\end{equation}
where $H$ and $W$  represent height, width  of the images $X$ and $X^{\prime}$. 
\subsubsection{Total-Variation loss term.}
To further improve the smoothness of the segmentation masks, a Total-Variation (TV) loss \cite{javanmardi_unsupervised_2018} is adopted and computed directly from the difference maps. TV loss enforces smoothness for the generated counterfactuals suppressing the noise and preserving the edges at the same time.  The formula for the objective function is as follows:

\begin{equation}
L_{tv} = \frac{1}{HW}\left(\sum_{i=1}^{H-1}\sum_{j=1}^{W} (x_{i+1,j} - x_{i,j})^2 + \sum_{i=1}^{H}\sum_{j=1}^{W - 1} (x_{i,j+1} - x_{i,j})^2\right),
\end{equation}
where $x_{i,j}$ is the intensity of a pixel at position $(i, j)$ in the input image $X$.

TV loss serves as a regularization term and improves consistency in the raw difference maps enforcing the model to perturb only densely located regions. 

\section{Experiments}
\subsection{Datasets}
\subsubsection{TotalSegmentator.}
The TotalSegmentator \cite{wasserthal_totalsegmentator_2023} dataset is an extensive collection of CT imaging data, particularly designed to train and evaluate algorithms for the task of image segmentation. Originating from a wide array of sources, it encompasses a diverse set of medical scans, including those of the chest, abdomen, and pelvis regions, among others. In addition to manually labelled ground truth masks, a great portion of the dataset is annotated with pre-trained segmentation models, which introduces some level of noise to the annotations. 

Synthetic anomalies are generated inside the kidneys of these scans for development of the model before moving to segmentation of real tumors. Generation of sythentic anomalies is described in \ref{app:synth_gen}. The dataset's scans are split randomly into training and validation sets with a ratio of 80\%/20\%.

\subsubsection{TUH.} The Tartu University Hospital kidney tumor dataset contains contrast enhanced CT scans of 291 kidney tumor cases and  300 control cases with pixel-level annotations for classes \textit{kidney, malignant lesion} and \textit{benign lesion}. Dataset was annotated by five radiologist and each scan was annotated by at least two of them. Final labels were produced by combining the two versions. If any major disagreements presented themselves they were resolved in direct discussion between radiologists. From pixel-level labels, the image-level labels are extracted of whether the slice contains a malignant lesion (kidney tumor) or not and used these labels for training of the classifier. Pixel-level labels were only used for evaluating the final segmentations quality. The dataset's scans are split randomly into training and validation sets  with a ratio of 80\%/20\% stratified by the total tumor area in voxels present in each scan.

\subsection{Evaluation}
\subsubsection{Realism of generated images.}
The Fréchet inception distance (FID) score \cite{heusel_gans_2018} is a widely used for measuring the similarity between generated images and a real sets of images. It involves the use of a pre-trained deep learning model to extract feature vectors from both sets of images. These features encapsulate various aspects of the images, such as textures, edges, and patterns. The similarity is defined as the distance between the activation distributions of the real image $x$ and the synthetic explanations $x_{c}$ as, 
\begin{equation}
FID(x, x_{\delta}) = ||\mu_{\bf x}-\mu_{\bf x_{\delta}}||_{2}^{2}+\mathrm{Tr}(\Sigma_{\bf x}+\Sigma_{\bf x_{\delta}}-2\big(\Sigma_{\bf x}\Sigma_{\bf x_{\delta}}\big)^{\frac{1}{2}}),
\end{equation}
where $\mu$’s and $\Sigma$’s are mean and covariance of the activation vectors derived from the penultimate layer of a pre-trained Inception v3 network.

\subsubsection{Classifier consistency.}

Counterfactual Validity (CV) score is a metric defines the fraction of counterfactual explanations that successfully flipped the classification decision, e.g if the input image is positive, the explanation should be predicted as negative. Prediction flip is considered successful when the difference of predictions $|f(X) - f(X_{cf})| $ is over a certain threshold $\tau$. Similarly to the original study \cite{singla_explaining_2022}, $\tau = 0.8$ is used.

\subsubsection{Segmentation metric.}
Intersection Over Union (IoU) score is a widely adopted metric for evaluating segmentation accuracy. It is computed as follows:
\begin{equation}
IoU (S, S_c)=\frac{|S \cap S_c|}{|S \cup S_c|}=\frac{T P}{T P+F P+F N},
\end{equation}
where $S$ and $S_c$ are ground truth and predicted segmentation masks respectively and $TP$, $FP$, $FN$ stand for true positive, false positve and false negative predictions.

\subsection{Implementation details}
All neural networks were implemented in PyTorch \cite{paszke_pytorch_2019} and were trained on the High Performance Computing (HPC) cluster of the University of Tartu on a single Nvidia A100-SXM-40GB GPU. Only kidney slices with kidney mask area of at least 32 pixels per image are used from each dataset, which are resized to 256x256 with bilinear interpolation and sampled into batches of 16 images. The training pipeline consists of two independent stages, namely classifier and explanation model training. The parameters of the classifier remain frozen throughout training process of the GAN. As for the classifier architecture, the models like ResNet18 \cite{he_deep_2015} for the synthetic anomalies and EfficientNet-V2 \cite{tan_efficientnetv2_2021} for real tumors are used respectively. The Adam \cite{kingma_adam_2017} optimizer is used for optimizing the objective function with parameters $\alpha=0.0002$, $\beta_1=0$, $\beta_2=0.9$. The segmentation masks from counterfactual generation are extracted as taking the absolute difference between the counterfactual and input images, and thresholding it with a fixed value. 

\subsection{Comparison with modified Singla et al.* method}
To establish a baseline comparison, the main purpose is to contrast the proposed method with that of Singla et al. However, the direct comparison is challenging due to the absence of the model's code, necessitating a unique implementation. To establish fair comparison, the reliance on segmentation masks should additionally be removed in the loss function, as the goal is to detect the masks for WSSS without using them explicitly. Thus, the loss terms  were adapted accordingly. The reconstruction loss is computed as a plain $L_1$ function averaging over all the pixels in the generated and input images. This modification, however, resulted in poor segmentation performance. Suspecting unreported skip connections in their cGAN model, the baseline was  enhanced by incorporating skip connections. In the given results, this adjusted model is referred as the modified Singla et al*.
  
\section{Results}
\begin{figure}[!htbp]
    \centering
    \includegraphics[width=1\linewidth]{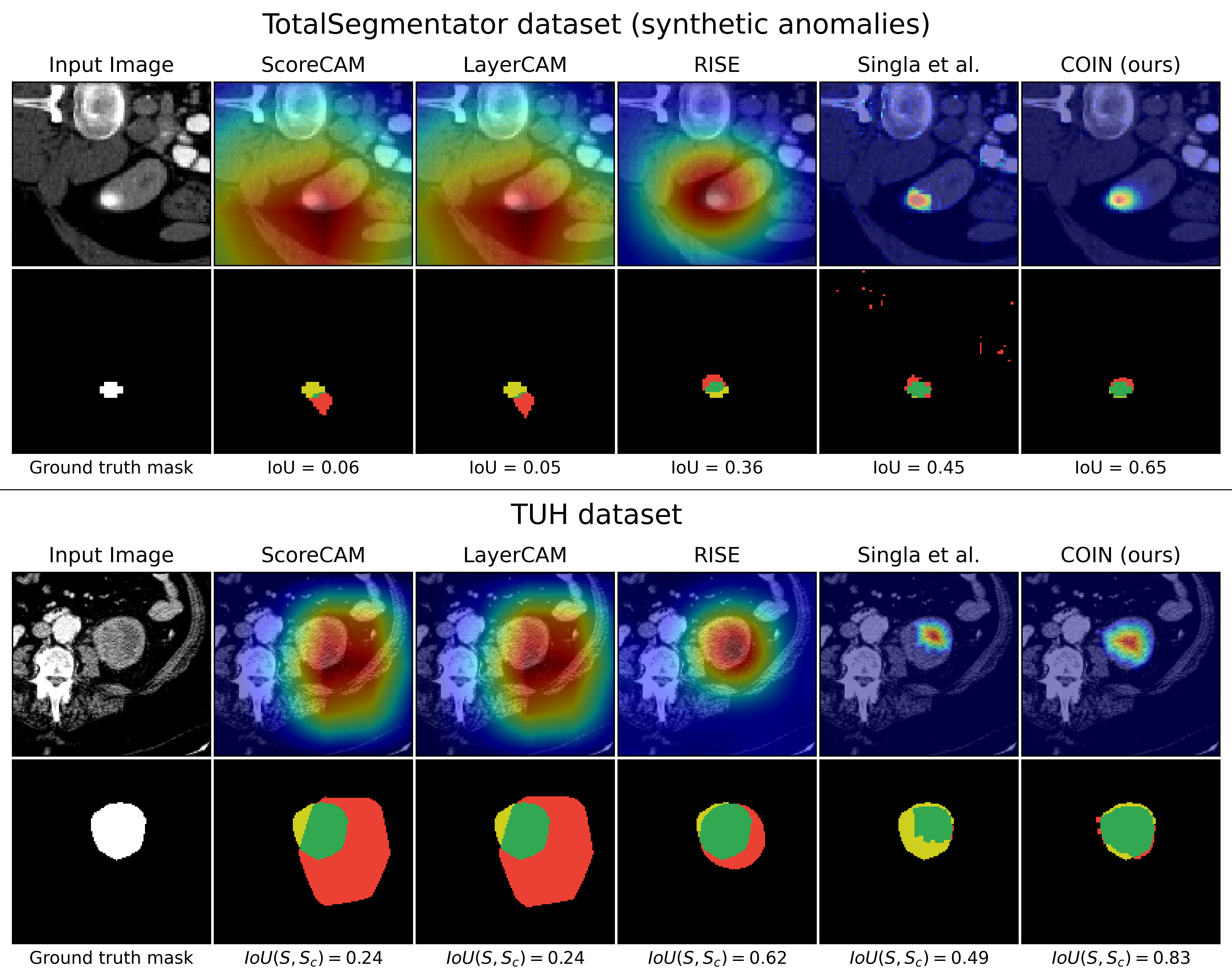}
    \caption{Visualization of the attribution and the proposed counterfactual inpainting pipeline methods' predictions on TotalSegmentator and TUH datasets. For each dataset, the bottom row depicts thresholded masks obtained from saliency maps from each method. For each masks, colors represent outcomes in terms of true positive (green), false positive (red) and false negative (yellow) predictions. White masks denote ground truth labels. Images are zoomed in for better clarity. }
    \label{fig:cf-inp-vs-attrib-tuh-vis}
\end{figure}

In this chapter, evaluation results are given for the proposed COIN method compared to attribution methods  \cite{jiang_layercam_2021,petsiuk_rise_2018,wang_score-cam_2020}. To convert the attention maps produced by all the methods, the outputs are normalized to the $[0; 1]$ range and threshold with a fixed value. The 0-1 range sweep is performed to find the best binarization threshold that maximizes IoU for each method. For the counterfactual methods, a morphological postprocessing of closing and opening is applied to suppress noise and retain only one largest connected component in the masks.  The results are presented in the Table \ref{cf-inp-vs-attrib-all-results} and the Figure~\ref{fig:cf-inp-vs-attrib-tuh-vis} for synthetic anomalies and real tumors respectively on a test set. In all the comparisons, COIN outperforms the alternative methods by a large margin in terms of IoU of up to 60\% on synthetic anomalies, and up to 14\% on real tumors. Moreover, while attribution methods like CAM and RISE are computationally inexpensive, as they do not require training an auxiliary model, they generally yield less accurate results and do not have static inference times, requiring multiple forward passes to compute saliency maps, which makes them less efficient for large datasets. In contrast, COIN, despite requiring approximately 20 hours of training on a single GPU, necessitates only a single forward pass and significantly enhances performance, achieving superior results that justify the additional computation time and making it more efficient for extensive data applications.

\begin{table}[!ht]
\centering
\caption{Metric results for the attribution methods and the proposed counterfactual inpainting pipeline on TUH dataset. Since CAMs and RISE do not create counterfactual images, FID and CV metrics cannot be computed for these methods.}
\begin{tabular}{clccc}
\toprule
\textbf{Datasets}& \textbf{Methods} & \textbf{FID}$\;\downarrow$ & \textbf{CV}$\;\uparrow$ & \textbf{IoU}$\;\uparrow$ \\
\midrule
& ScoreCAM    & -  & - & 0.030
\\
& LayerCAM     & - & - & 0.026
\\
\multirow{1}{*}{TotalSegmentator}& RISE     & - & -  & 0.397
\\
& Singla et al.*     & 0.047 & 0.998  & 0.445
\\
& \textbf{COIN}    & \textbf{0.003} & \textbf{0.997}  & \textbf{0.646}\\
\midrule
& ScoreCAM    & -  & - & 0.293
\\
& LayerCAM     & - & - & 0.296
\\
\multirow{1}{*}{Tartu University Hospital\;\;}& RISE     & - & -  & 0.294
\\
& Singla et al.*     & 0.203 & 0.992  & 0.352
\\
& \textbf{COIN}    & \textbf{0.036} & \textbf{0.980}  & \textbf{0.432}\\

\bottomrule
\end{tabular}
\label{cf-inp-vs-attrib-all-results}
\end{table}

\subsection{Ablation experiments}
In the ablation study of this research, the importance of each loss term's contribution to the final image realism is assessed, classifier predictions flip rate and segmentation accuracy for the proposed counterfactual inpainting pipeline. This was achieved by zeroing out the weights of each loss term independently and compare the metric results.  The Table \ref{loss-abl-study} presents the results of the experiment.

\begin{table}
\centering
\caption{Ablation study results on each loss term contribution based on TotalSegmentator dataset with synthetic anomalies.}
\begin{tabular}{clcc}
\toprule
\textbf{Experiment} & \textbf{FID}$\;\downarrow$ & \textbf{CV}$\;\uparrow$ & \textbf{IoU}$\;\uparrow$ \\
\midrule
$\lambda_{idt}=0$   &  0.0024& 0.997&0.500\\
$\lambda_{f}=0$    & 0.0032& 0.642& 0.424     \\
$\lambda_{tv}=0$   & 0.0178& 0.997& 0.427  \\
\midrule
COIN baseline    & \textbf{0.0029}& \textbf{0.997}& \textbf{0.646}  \\

\bottomrule
\end{tabular}
\label{loss-abl-study}
\end{table}

Classifier consistency loss plays a crucial role for learning good counterfactual images, since excluding it from the loss practically removes the influence of the classifier on the generator outputs, resulting in CV score degradation by 35\%. Self-Consistency and TV losses are important for the model to avoid random perturbations and force the model to focus on perturbing only densely located regions, which enforces minimum change and increases IoU by up to 22\%. 

\section{Discussion}
In this study, a novel counterfactual inpainting approach is introduced for weakly supervised semantic segmentation, which demonstrated to outperform existing attribution methods and the baseline counterfactual method in the segmentation of synthetic anomalies and real tumors on Tartu University Hospital dataset. Aiming for a fair comparison, the best performing threshold is meticulously selected for all attribution methods.  Still, CAM methods substantially underperformed compared to the proposed approach on both synthetic anomalies and real tumors. They primarily rely on high-level classification-relevant image areas and often fail to localize the central parts of anomalies accurately. RISE proved to be more effective than ScoreCAM and LayerCAM methods as it produces the saliency map based on the direct perturbations of the input image with pre-generated masks. The intentions for comparison with original Singla et al. approach were challenged by the unavailability of the code. This is addressed through a custom implementation, enhancing their method with added skip connections. It is assumed that the original methodology might have underreported architectural details, given its poor initial performance. The modified Singla et al.* approach achieved high segmentation results after postprocessing but produced lower fidelity images resulting in high FID score. Overall, COIN generates consistent difference maps, enhancing the accuracy and confirming suitability for the WSSS task. Figures~\ref{fig:cf-inp-vs-attrib-tuh-vis} offers a qualitative evaluation of the generated counterfactuals for the discussed methods.
\subsubsection{Limitations and Future works.} 
A key limitation of the effectiveness of the developed counterfactual inpainting pipeline is its dependency on the performance of the underlying black-box classifier. Although simpler than training a segmentation model, training a classifier to a satisfactory level of accuracy and robustness may become a non-trivial task that requires substantial amounts of labeled data, computational resources, and careful tuning of model parameters. Any imperfections in the classifier, such as biases in the training data, overfitting, or underfitting, can adversely affect the quality of the generated counterfactuals. This, in turn, can lead to poor segmentation labels, which may not accurately reflect the underlying data distribution or the specific features that are of interest in the analysis. 

Another notable limitation of the current pipeline is its restriction to 2D analysis, despite the inherently 3D nature of the CT scans dataset. This simplification can lead to a loss of spatial context and information that is crucial for accurate segmentation and analysis of medical images. This limitation highlights the need for extending the pipeline to accommodate 3D data directly which is planned for the future works. Developing methods that can efficiently process and generate counterfactuals in 3D would significantly enhance the applicability and effectiveness of the pipeline in medical imaging contexts.
Recognizing this, future work will focus on extending COIN to accommodate 3D data, thereby enhancing the precision of segmentation masks as 3D input will provide a more comprehensive understanding of the input.
Additionally, the weakly supervised segmentation pipeline with counterfactual inpainting should not be confined to tumor data and medical domain in general. Future studies will assess the generality of COIN method, testing its effectiveness across diverse applications and datasets. This exploration is anticipated to be particularly beneficial in scenarios where acquiring segmentation masks is more challenging than obtaining classification labels.

\section{Conclusion}
In this study, a novel strategy is proposed that utilizes explainability for weakly supervised semantic segmentation task for medical domain. By adopting a counterfactual method as a foundation, the method is enhanced with a perturbation-based generator, simplified conditioning for inpainting or removing abnormality, elimination of the need for segmentation masks, and a new loss term for enforcing smoothness in the counterfactuals. All these additions contributed to precise generation of segmentation masks, demonstrating superiority over attribution methods and original counterfactual approach. This innovative approach enhances the capability to generate more nuanced and detailed counterfactual examples resulting in a significant contribution to the field of weakly supervised learning. By addressing the inherent limitations of sparse annotations and leveraging the power of counterfactual reasoning, the inpainting pipeline offers a robust solution for improving semantic segmentation models without the need for extensive manually labeled datasets. The code is released in the public repository: 
\begin{verbatim}
    https://github.com/Dmytro-Shvetsov/counterfactual-search
\end{verbatim}

\begin{credits}
\subsubsection{\ackname}
The data has been collected as part of the Clinical Investigation that has been approved by The University of Tartu Research Ethics Committee (no 332/T-9, dated 21.12.2020), we thank Better Medicine for providing the dataset.
This research was supported by the Estonian Research Council Grants PRG1604, the European Union’s Horizon 2020 Research and Innovation Programme under Grant Agreement No. 952060 (Trust AI), by the Estonian Centre of Excellence in Artificial Intelligence (EXAI), funded by the Estonian Ministry of Education and Research, by Better Medicine and by STACC. 
\subsubsection{\discintname}
Dmytro Fishman owns stock in Better Medicine. All other authors have no competing interests to declare that are
relevant to the content of this article. 
\end{credits}

\newpage
\section*{Appendix A}

In this chapter, a detailed experimentation is given for the iterative improvements to the Singla et al.* method to obtain the COIN pipeline. Table \ref{main-synth-results} summarizes all the performed experiments and obtained metrics based on the TotalSegmentator and synthetic anomalies.

\begin{table}[]
\centering
\caption{Metric results for the iterative improvements of the original Singla et al. and the COIN methods. Experiment H refers to the modified Singla et al.* method.}\label{tab:iterative_impr}
\begin{tabular}{l|p{0.5cm}p{0.5cm}p{0.5cm}p{1cm}ccc}
\multicolumn{1}{c}{ID \& iterative changes}&\rotatebox[origin=l]{45}{uses masks}&\rotatebox[origin=l]{45}{perturbations} &\rotatebox[origin=l]{45}{skip connections} &\rotatebox[origin=l]{45}{conditions} & \textbf{FID}$\;\downarrow$ & \textbf{CV}$\;\uparrow$ & \textbf{IoU}$\;\uparrow$ \\
\midrule
A&\;\checkmark&           &0&2 &  1.6190& 0.992&0.020
\\
B (+ perturbations)&\;\checkmark&\checkmark &0&2   & \textbf{0.5500}& \textbf{0.934}& \textbf{0.086}
\\
C (+ skip connection)&\;\checkmark&\checkmark &1&2   & 0.3254& 0.968& 0.284
\\
D (+ skip connection)&\;\checkmark&\checkmark &2&2   & 0.1751& 0.933& 0.480
\\
E (+ skip connection)&\;\checkmark&\checkmark &3&2   & 0.0849& 0.961& 0.463
\\
 F (+ skip connection)& \;\checkmark& \checkmark & 4& 2   & \textbf{0.0253}& 0.9542&\textbf{0.509}
\\
 G (- masks)& & \checkmark & 4& 2   & 0.0493& 0.996&0.528
\\
 H (- perturbations)& & & 4& 2   & 0.0466& 0.998&0.445
\\
\midrule
COIN&&\checkmark &4&1  & \textbf{0.0029}& \textbf{0.997}& \textbf{0.646}\\

\bottomrule
\end{tabular}
\label{main-synth-results}
\end{table}

\subsection{Loss function for dual-conditioning in Singla et al.* }  
In this chapter, the main difference between COIN and the original Singla et al.* is described in terms of  loss functions for training the image generation model. Firstly, the model of Singla et al.* with two conditions accepts an input image $X$ and a condition parameter $\delta$, so that the counterfactual example $X_{cf} = \mathscr{E}(X, 1 - f(X))$ yields a \textbf{normal} image if $X$ is \textbf{abnormal} or an \textbf{abnormal} image if $X$ is \textbf{normal}. To achieve this, the \textbf{classification model consistency loss} is introduced as follows:
\begin{equation}
{\mathcal L}_{f} = D_{\mathrm{KL}}(f(X_{cf})\ ||\ 1 - f(X)),
\end{equation}
In terms of \textbf{domain-aware self-consistency loss}, to achieve corresponding counterfactual images when applying a series generations, the objective functions is given as follows:
\begin{equation}
{\mathcal{L}}_{\mathrm{idt}}={\mathcal{L}}_{\mathrm{rec}}(X,\mathscr{E}(X, f(X)))\ +\ {\mathcal{L}}_{\mathrm{rec}}(X,\mathscr{E}(X, 1 - f(X)), f(X)),
\end{equation}
where both components enforce the model to predict the identity image. However, the first component should generate one if conditioned on the $f(X)$, whereas the second one should generate the identity when doing counterfactual generation two times in a row.  For the experiments where it is referred that segmentation masks are used, the ${\mathcal{L}}_{\mathrm{rec}}$ function is defined as the average L1 distance computed between foreground pixels of segmentation mask $S$ for label $j$.
\begin{equation}
{\mathcal{L}}_{\mathrm{rec}}(X,X^{\prime})=\sum_{j}{\frac{S_{j}(X)\cdot||X-X^{\prime}||_1.}{\sum S_{j}(X)}},\end{equation}
In the case of the experiment with no masks used, ${\mathcal{L}}_{\mathrm{rec}}$ is the $L_1$ function similar to COIN.

\subsection{Synthetic anomaly generation}\label{app:synth_gen}
\begin{figure}[!ht]
    \centering
    \includegraphics[width=1\linewidth]{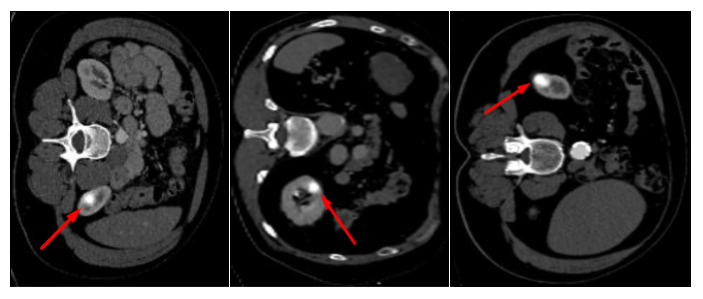}
    \caption{Examples of the synthetic anomalies injected randomly inside kidneys for the TotalSegmentator dataset.}
    \label{fig:synth_examples}
\end{figure}
In order to establish a robust benchmark for assessing the performance of the modeling decisions, this research introduces a synthetic anomaly, meticulously designed and integrated into the CT scans datasets. The synthetic anomaly is conceptualized as a Gaussian blob, characterized by a fixed sigma and radius. This design choice is deliberate, aiming to mimic typical radiological findings that present as circular or ellipsoid structures in medical imaging.  The synthetic anomaly is sampled at random positions within one of the kidneys in the abdominal slices of the scans in the TotalSegmentator dataset. This approach ensures a diverse and unpredictable distribution of anomalies, closely simulating the randomness and variability. To further enhance the complexity and variability of the synthetic anomaly, a series of transformation and augmentation techniques are employed, including random grid distortions, scaling and rotations. The examples of resulting gaussian blobs are visualized in Figure \ref{fig:synth_examples}.

\subsection{Original vs Perturbation-based generator}
\begin{figure}[!ht]
    \centering
    \includegraphics[width=1\linewidth]{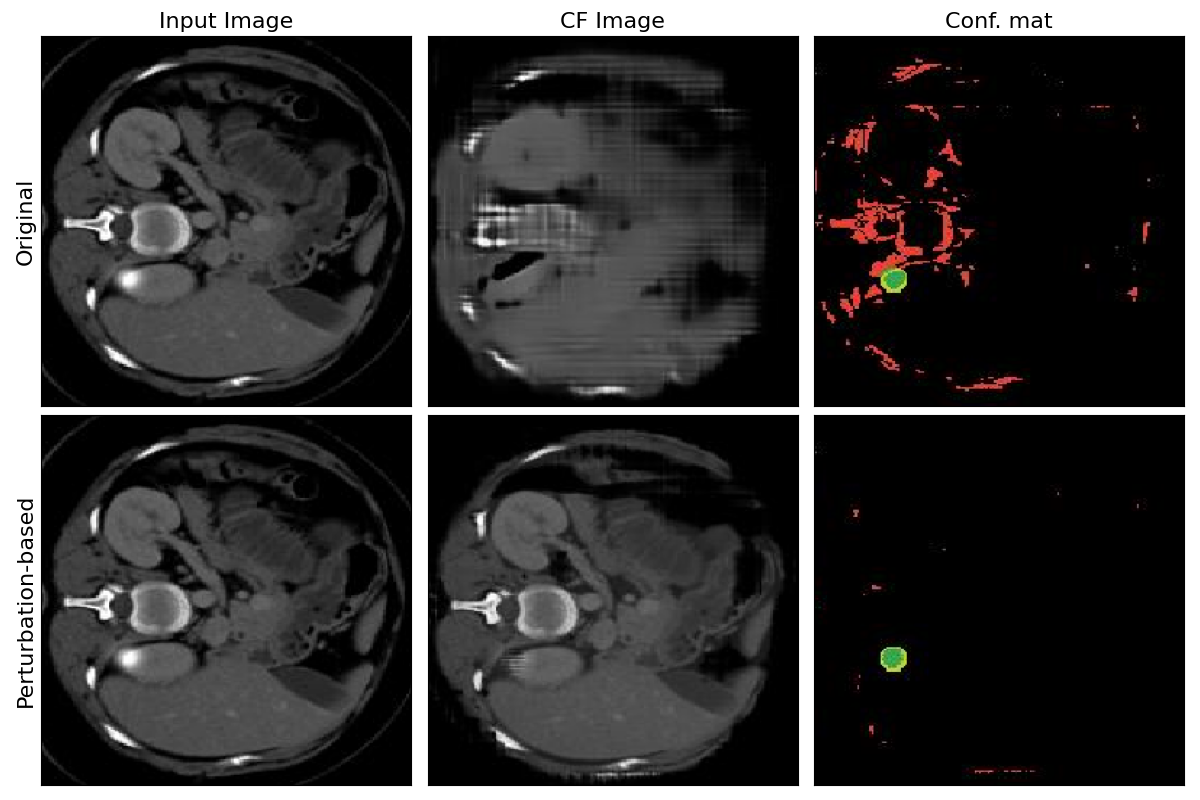}
    \caption{Examples of images generated with original and perturbation-based Singla et al.*  pipelines.}
    \label{fig:orig-vs-ptb-based-vis}
\end{figure}

Within this experiment, the counterfactual explainer of Singla et al. is taken to validate the significant improvement employing the perturbation-based generation in terms of FID and IoU scores. The perturbation-based image generation generates much higher fidelity images. Instead of reconstructing the whole input image from scratch, the decoder learns to output only the changes needed to flip classifier decision. Figure \ref{fig:orig-vs-ptb-based-vis} gives qualitative evaluation of the generated images following the two approaches.

\subsection{Influence of skip connections on the generated images quality}
\begin{figure}[!ht]
    \centering
    \includegraphics[width=1\linewidth]{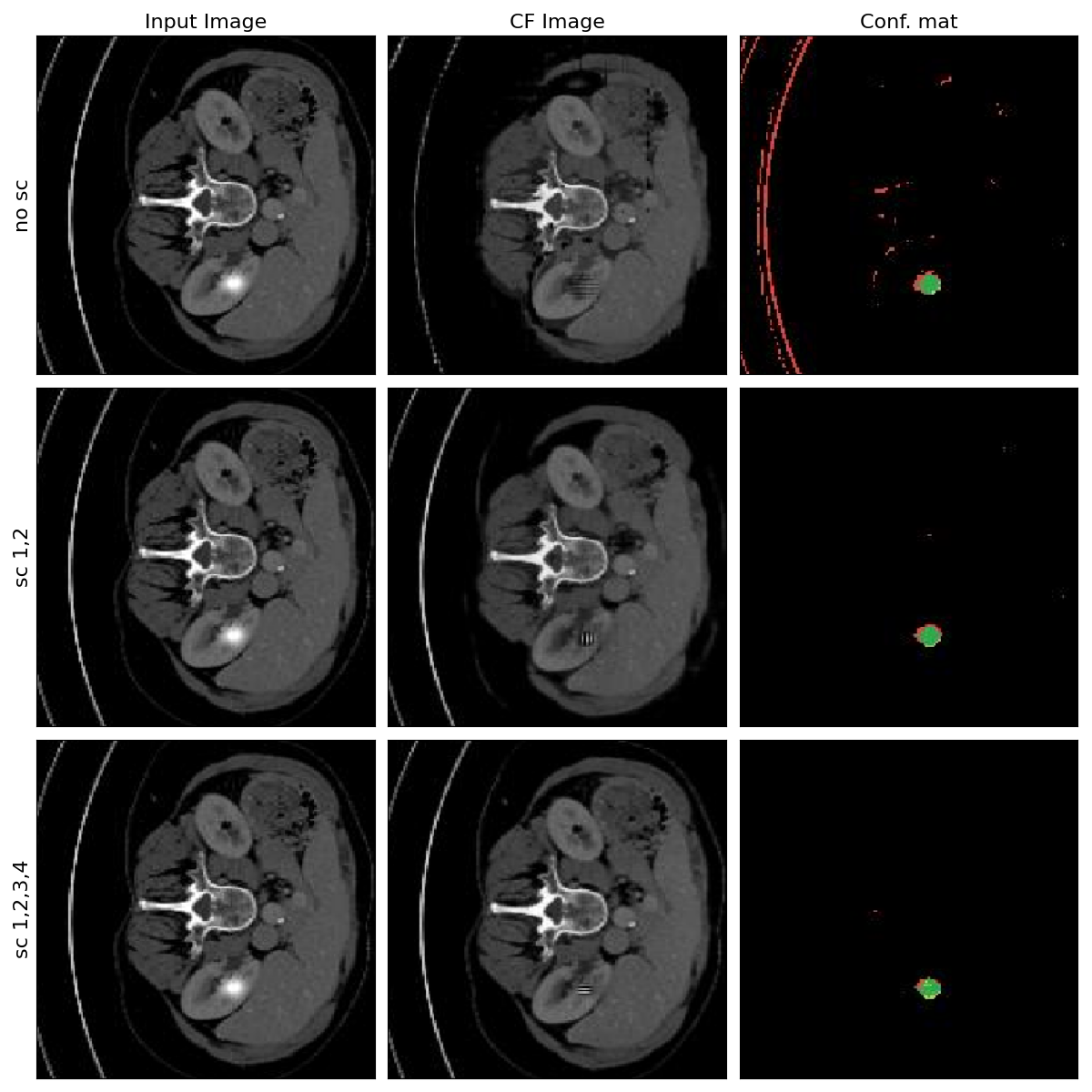}
    \caption{Examples of images generated with and without skip-connections between encoder-decoder layers of the perturbation-based Singla et al.* pipeline.}
    \label{fig:skip-conn-vis}
\end{figure}
In this experiment, the baseline of Singla et al. employing perturbation-based counterfactual generation from the previous section is taken to showcase the importance of skip connections in the generator network to mitigate drastic distortions of the input images. During the down-sampling process of the encoder, the information loss is inevitable, so reconstructing the counterfactual images with minimum perturbations becomes a challenge. Therefore, the skip connections between different down-sampling and up-sampling layers are gradually injected to show the improvements in terms of FID and IoU scores.  The perturbation-based image generation leveraging skip connections results in less distorted images, hence, in lower FID score. Figure \ref{fig:skip-conn-vis} gives qualitative evaluation of the generated images with and without adoption of skip-connections.

\subsection{Counterfactual Explanation vs Counterfactual Inpainting Segmentation Accuracy}
\begin{figure}[!ht]
    \centering
    \includegraphics[width=1\linewidth]{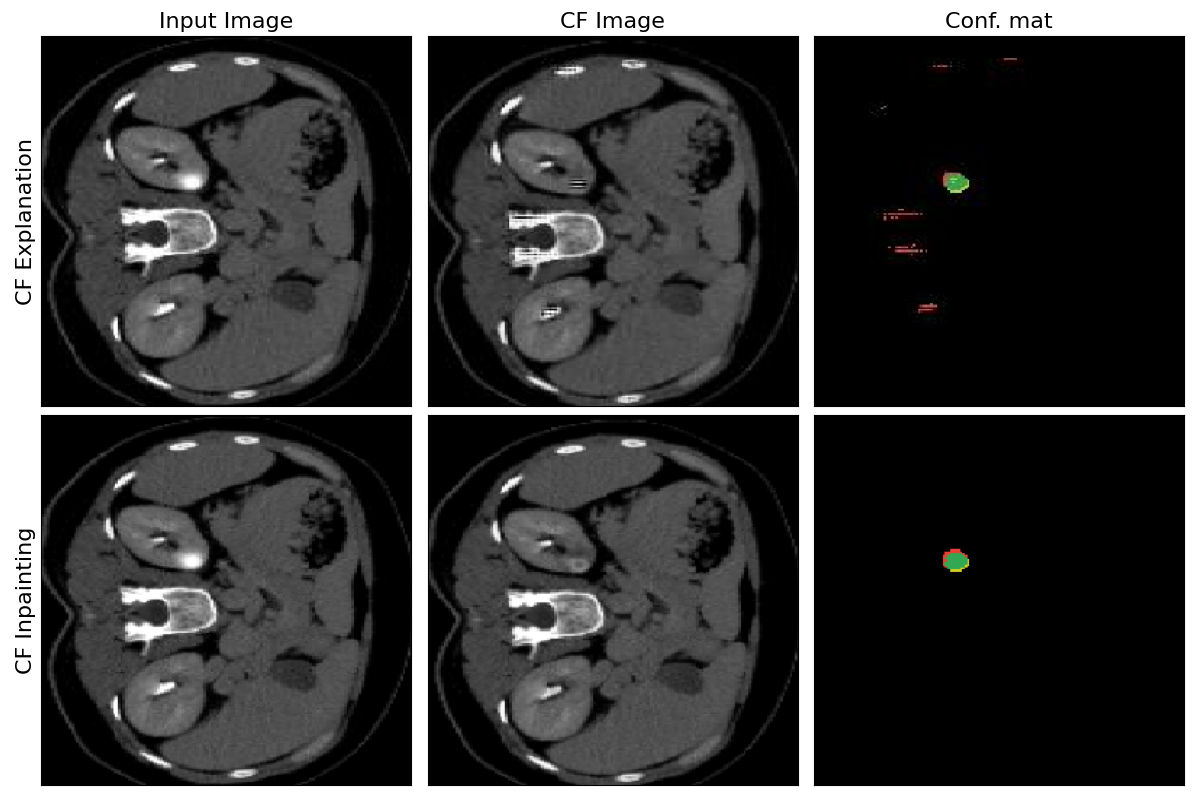}
    \caption{Examples of images generated with perturbation-based Singla et al.* method equipped with skip connections and with the proposed counterfactual inpainting approach.}
    \label{fig:base-cf-vs-cf-inp-vis}
\end{figure}
This experiment proves that the proposed counterfactual inpainting pipeline outperforms the base counterfactual explanation approach. Both methods are trained and evaluated in terms of segmentation accuracy for the extracted weak segmentation labels from the counterfactual images. 

The benefits of using the counterfactual inpainting are two-fold. First, it does not require segmentation masks for enforcing local consistency. Second, the IoU score is much higher due to the fact that the model is simplified to only either inpaint the anomaly or not to produce the segmentation mask. Figure \ref{fig:base-cf-vs-cf-inp-vis} gives qualitative evaluation of the generated counterfactuals following the two approaches.

\bibliographystyle{splncs04}

\begin{thebibliography}{10}
\providecommand{\url}[1]{\texttt{#1}}
\providecommand{\urlprefix}{URL }
\providecommand{\doi}[1]{https://doi.org/#1}

\bibitem{ahn2018learning}
Ahn, J., Kwak, S.: Learning pixel-level semantic affinity with image-level supervision for weakly supervised semantic segmentation. In: Proceedings of the IEEE conference on computer vision and pattern recognition. pp. 4981--4990 (2018)

\bibitem{akula2022cx}
Akula, A.R., Wang, K., Liu, C., Saba-Sadiya, S., Lu, H., Todorovic, S., Chai, J., Zhu, S.C.: Cx-tom: Counterfactual explanations with theory-of-mind for enhancing human trust in image recognition models. Iscience  \textbf{25}(1) (2022)

\bibitem{atad2022chexplaining}
Atad, M., Dmytrenko, V., Li, Y., Zhang, X., Keicher, M., Kirschke, J., Wiestler, B., Khakzar, A., Navab, N.: Chexplaining in style: Counterfactual explanations for chest x-rays using stylegan. arXiv preprint arXiv:2207.07553  (2022)

\bibitem{bischof_counterfactual_2023}
Bischof, R., Scheidegger, F., Kraus, M.A., Malossi, A.C.I.: Counterfactual image generation for adversarially robust and interpretable classifiers, \url{http://arxiv.org/abs/2310.00761}

\bibitem{burton2019using}
Burton, R.J., Albur, M., Eberl, M., Cuff, S.M.: Using artificial intelligence to reduce diagnostic workload without compromising detection of urinary tract infections. BMC medical informatics and decision making  \textbf{19},  1--11 (2019)

\bibitem{byrne2019counterfactuals}
Byrne, R.M.: Counterfactuals in explainable artificial intelligence (xai): Evidence from human reasoning. In: IJCAI. pp. 6276--6282 (2019)

\bibitem{chaddad2023survey}
Chaddad, A., Peng, J., Xu, J., Bouridane, A.: Survey of explainable ai techniques in healthcare. Sensors  \textbf{23}(2), ~634 (2023)

\bibitem{chen2020weakly}
Chen, L., Wu, W., Fu, C., Han, X., Zhang, Y.: Weakly supervised semantic segmentation with boundary exploration. In: Computer Vision--ECCV 2020: 16th European Conference, Glasgow, UK, August 23--28, 2020, Proceedings, Part XXVI 16. pp. 347--362. Springer (2020)

\bibitem{chen2022c}
Chen, Z., Tian, Z., Zhu, J., Li, C., Du, S.: C-cam: Causal cam for weakly supervised semantic segmentation on medical image. In: Proceedings of the IEEE/CVF Conference on Computer Vision and Pattern Recognition. pp. 11676--11685 (2022)

\bibitem{cui2020unified}
Cui, H., Wei, D., Ma, K., Gu, S., Zheng, Y.: A unified framework for generalized low-shot medical image segmentation with scarce data. IEEE Transactions on Medical Imaging  \textbf{40}(10),  2656--2671 (2020)

\bibitem{ghassemi2021false}
Ghassemi, M., Oakden-Rayner, L., Beam, A.L.: The false hope of current approaches to explainable artificial intelligence in health care. The Lancet Digital Health  \textbf{3}(11),  e745--e750 (2021)

\bibitem{gidde2021validation}
Gidde, P.S., Prasad, S.S., Singh, A.P., Bhatheja, N., Prakash, S., Singh, P., Saboo, A., Takhar, R., Gupta, S., Saurav, S., et~al.: Validation of expert system enhanced deep learning algorithm for automated screening for covid-pneumonia on chest x-rays. Scientific Reports  \textbf{11}(1),  23210 (2021)

\bibitem{guidotti2022counterfactual}
Guidotti, R.: Counterfactual explanations and how to find them: literature review and benchmarking. Data Mining and Knowledge Discovery pp. 1--55 (2022)

\bibitem{he_deep_2015}
He, K., Zhang, X., Ren, S., Sun, J.: Deep residual learning for image recognition, \url{http://arxiv.org/abs/1512.03385}

\bibitem{heusel_gans_2018}
Heusel, M., Ramsauer, H., Unterthiner, T., Nessler, B., Hochreiter, S.: {GANs} trained by a two time-scale update rule converge to a local nash equilibrium, \url{http://arxiv.org/abs/1706.08500}

\bibitem{javanmardi_unsupervised_2018}
Javanmardi, M., Sajjadi, M., Liu, T., Tasdizen, T.: Unsupervised total variation loss for semi-supervised deep learning of semantic segmentation, \url{http://arxiv.org/abs/1605.01368}

\bibitem{jeanneret_adversarial_2023}
Jeanneret, G., Simon, L., Jurie, F.: Adversarial counterfactual visual explanations. In: 2023 {IEEE}/{CVF} Conference on Computer Vision and Pattern Recognition ({CVPR}). pp. 16425--16435. {IEEE}. \doi{10.1109/CVPR52729.2023.01576}, \url{https://ieeexplore.ieee.org/document/10205255/}

\bibitem{jiang_layercam_2021}
Jiang, P.T., Zhang, C.B., Hou, Q., Cheng, M.M., Wei, Y.: {LayerCAM}: Exploring hierarchical class activation maps for localization  \textbf{30},  5875--5888. \doi{10.1109/TIP.2021.3089943}, \url{https://ieeexplore.ieee.org/document/9462463/}

\bibitem{karimi2022survey}
Karimi, A.H., Barthe, G., Sch{\"o}lkopf, B., Valera, I.: A survey of algorithmic recourse: contrastive explanations and consequential recommendations. ACM Computing Surveys  \textbf{55}(5),  1--29 (2022)

\bibitem{keil2006explanation}
Keil, F.C.: Explanation and understanding. Annu. Rev. Psychol.  \textbf{57},  227--254 (2006)

\bibitem{kenny2021generating}
Kenny, E.M., Keane, M.T.: On generating plausible counterfactual and semi-factual explanations for deep learning. In: Proceedings of the AAAI Conference on Artificial Intelligence. vol.~35, pp. 11575--11585 (2021)

\bibitem{kingma_adam_2017}
Kingma, D.P., Ba, J.: Adam: A method for stochastic optimization, \url{http://arxiv.org/abs/1412.6980}

\bibitem{miller2019explanation}
Miller, T.: Explanation in artificial intelligence: Insights from the social sciences. Artificial intelligence  \textbf{267},  1--38 (2019)

\bibitem{miller2021contrastive}
Miller, T.: Contrastive explanation: A structural-model approach. The Knowledge Engineering Review  \textbf{36}, ~e14 (2021)

\bibitem{miyato_spectral_2018}
Miyato, T., Kataoka, T., Koyama, M., Yoshida, Y.: Spectral normalization for generative adversarial networks, \url{http://arxiv.org/abs/1802.05957}

\bibitem{musen2021clinical}
Musen, M.A., Middleton, B., Greenes, R.A.: Clinical decision-support systems. In: Biomedical informatics: computer applications in health care and biomedicine, pp. 795--840. Springer (2021)

\bibitem{paszke_pytorch_2019}
Paszke, A., Gross, S., Massa, F., Lerer, A., Bradbury, J., Chanan, G., Killeen, T., Lin, Z., Gimelshein, N., Antiga, L., Desmaison, A., Köpf, A., Yang, E., {DeVito}, Z., Raison, M., Tejani, A., Chilamkurthy, S., Steiner, B., Fang, L., Bai, J., Chintala, S.: {PyTorch}: An imperative style, high-performance deep learning library, \url{http://arxiv.org/abs/1912.01703}

\bibitem{pearl2019seven}
Pearl, J.: The seven tools of causal inference, with reflections on machine learning. Communications of the ACM  \textbf{62}(3),  54--60 (2019)

\bibitem{petsiuk_rise_2018}
Petsiuk, V., Das, A., Saenko, K.: {RISE}: Randomized input sampling for explanation of black-box models, \url{http://arxiv.org/abs/1806.07421}

\bibitem{ronneberger_u-net_2015}
Ronneberger, O., Fischer, P., Brox, T.: U-net: Convolutional networks for biomedical image segmentation, \url{http://arxiv.org/abs/1505.04597}

\bibitem{selvaraju2017grad}
Selvaraju, R.R., Cogswell, M., Das, A., Vedantam, R., Parikh, D., Batra, D.: Grad-cam: Visual explanations from deep networks via gradient-based localization. In: Proceedings of the IEEE international conference on computer vision. pp. 618--626 (2017)

\bibitem{shen2023survey}
Shen, W., Peng, Z., Wang, X., Wang, H., Cen, J., Jiang, D., Xie, L., Yang, X., Tian, Q.: A survey on label-efficient deep image segmentation: Bridging the gap between weak supervision and dense prediction. IEEE Transactions on Pattern Analysis and Machine Intelligence  (2023)

\bibitem{singla_explaining_2022}
Singla, S., Eslami, M., Pollack, B., Wallace, S., Batmanghelich, K.: Explaining the black-box smoothly- a counterfactual approach, \url{http://arxiv.org/abs/2101.04230}

\bibitem{tajbakhsh2020embracing}
Tajbakhsh, N., Jeyaseelan, L., Li, Q., Chiang, J.N., Wu, Z., Ding, X.: Embracing imperfect datasets: A review of deep learning solutions for medical image segmentation. Medical image analysis  \textbf{63},  101693 (2020)

\bibitem{tan_efficientnetv2_2021}
Tan, M., Le, Q.V.: {EfficientNetV}2: Smaller models and faster training, \url{http://arxiv.org/abs/2104.00298}

\bibitem{wachter2017counterfactual}
Wachter, S., Mittelstadt, B., Russell, C.: Counterfactual explanations without opening the black box: Automated decisions and the gdpr. Harv. JL \& Tech.  \textbf{31}, ~841 (2017)

\bibitem{wang_score-cam_2020}
Wang, H., Wang, Z., Du, M., Yang, F., Zhang, Z., Ding, S., Mardziel, P., Hu, X.: Score-{CAM}: Score-weighted visual explanations for convolutional neural networks, \url{http://arxiv.org/abs/1910.01279}

\bibitem{wasserthal_totalsegmentator_2023}
Wasserthal, J., Breit, H.C., Meyer, M.T., Pradella, M., Hinck, D., Sauter, A.W., Heye, T., Boll, D., Cyriac, J., Yang, S., Bach, M., Segeroth, M.: {TotalSegmentator}: robust segmentation of 104 anatomical structures in {CT} images  \textbf{5}(5),  e230024. \doi{10.1148/ryai.230024}, \url{http://arxiv.org/abs/2208.05868}

\bibitem{zemni_octet_2023}
Zemni, M., Chen, M., Zablocki, E., Ben-Younes, H., Perez, P., Cord, M.: {OCTET}: Object-aware counterfactual explanations. In: 2023 {IEEE}/{CVF} Conference on Computer Vision and Pattern Recognition ({CVPR}). pp. 15062--15071. {IEEE}. \doi{10.1109/CVPR52729.2023.01446}, \url{https://ieeexplore.ieee.org/document/10205035/}

\end{thebibliography}

\end{document}